\RequirePackage{amsmath}
\documentclass[runningheads]{llncs}
\usepackage[T1]{fontenc}
\usepackage{graphicx}
\usepackage{xcolor}  
\usepackage{bm,amssymb,bbold}
\usepackage{booktabs}
\usepackage{multirow}
\usepackage{array}
\usepackage[misc]{ifsym}
\usepackage{subcaption}

\DeclareMathOperator*{\argmax}{arg\,max}
\DeclareMathOperator*{\Min}{min}
\DeclareMathOperator*{\Max}{max}
\DeclareMathOperator*{\softmax}{softmax}
\DeclareMathOperator*{\Gumbel}{Gumbel}
\DeclareMathOperator*{\Value}{value}
\DeclareMathOperator*{\encode}{encode}
 
\newcommand{\pluseq}{\mathrel{+}=}
\newcommand{\subeq}{\mathrel{-}=}
\newtheorem{prop}[theorem]{Proposition}
\usepackage{algorithm}
\usepackage{algpseudocode}
\usepackage{enumitem}  

\newcounter{MyCounter}

\newcommand*{\MyToprule}{%
  \cmidrule[\heavyrulewidth]%
}
\newcommand*{\MyBottomrule}{%
  \cmidrule[\heavyrulewidth]%
}

\begin{document}
\title{Deep Explainable Relational Reinforcement Learning: A Neuro-Symbolic Approach}

\author{Rishi Hazra\inst{1} \Letter \and Luc De Raedt\inst{1,2}}
\institute{Centre for Applied Autonomous Sensor Systems (AASS), \\Örebro University, Sweden \and Department of Computer Science, KU Leuven, Belgium\\\email{\{rishi.hazra,luc.de-raedt\}@oru.se}}
\tocauthor{Rishi Hazra, Luc De Raedt}
\toctitle{Deep Explainable Relational Reinforcement Learning: A Neuro-Symbolic Approach}
\maketitle              
\begin{abstract}
Despite numerous successes in Deep Reinforcement Learning (DRL), the learned policies are not interpretable. Moreover, since DRL does not exploit symbolic relational representations, it has difficulties in coping with structural changes in its environment (such as increasing the number of objects). Relational Reinforcement Learning, on the other hand, inherits the relational representations from symbolic planning to learn reusable policies. However, it has so far been unable to scale up and exploit the power of deep neural networks. We propose Deep Explainable Relational Reinforcement Learning (DERRL), a framework that exploits the best of both -- \textit{neural} and \textit{symbolic} worlds. By resorting to a neuro-symbolic approach, DERRL combines relational representations and constraints from symbolic planning with deep learning to extract interpretable policies. These policies are in the form of logical rules that explain how each decision (or action) is arrived at. Through several experiments, in setups like the Countdown Game, Blocks World, Gridworld, and Traffic, we show that the policies learned by DERRL can be applied to different configurations and contexts, hence generalizing to environmental modifications\let\thefootnote\relax\footnotetext{\textcolor{blue}{Accepted in ECML PKDD 2023}}.

\end{abstract}
\section{Introduction}
\label{section:introduction}

Deep Reinforcement Learning (DRL)~\cite{8103164} has gained great success in many domains. However, so far, it has had limited success in relational domains, which are typically used in symbolic planning~\cite{10.5555/975615}. In the prototypical blocks world game (Figure 1), one goal is to place block $a$ on block $b$. An obvious plan for achieving this is to \textit{unstack} the blocks until both blocks $a$ and $b$ are at the top, upon which block $a$ can be moved atop block $b$. Standard DRL approaches struggle to adapt to out-of-domain data, such as  placing block $c$ on $d$, or applying the learned strategies to changes in the stack size or the number of stacks, thus failing to learn generalized policies. Furthermore, the black-box nature of the learned policies makes it difficult to interpret action choices, especially in domains involving transparency and safety~\cite{Lee2004TrustIA,Stowers2016IntelligentAT,10.1145/3278721.3278776}. Understanding a machine’s decision-making is crucial for human operators to eliminate irrational reasoning~\cite{10.1007/978-3-319-07458-0_24,DBLP:journals/corr/abs-1806-00069}. 

Relational Reinforcement Learning (RRL)~\cite{10.5555/645527.657460,Tadepalli04relationalreinforcement} combines symbolic planning and reinforcement learning and has origins in Statistical Relational AI and Inductive Logic Programming (ILP)~\cite{Muggleton94inductivelogic,DeRaedtKerstingEtAl16}. RRL uses logic programs to represent interpretable policies that are similar to symbolic planning languages~\cite{FIKES1971189,Ghallab98,gel98}. These policies use relations and objects, rather than specific states and actions, allowing agents to reason about their actions at a higher level of abstraction, and applying the learned knowledge to different situations. Earlier RRL approaches were purely symbolic~\cite{10.5555/645527.657460,driessens2004integrating,kersting2008non,martinez2017relational}, searching policy spaces guided by performance estimates, but did not exploit deep learning advancements and were not robust to noisy data. Recent approaches~\cite{DBLP:journals/corr/abs-1806-01830} use neural networks for scalability and improved internal representations, but learned policies are not human-readable.


\begin{figure*}[t]
	\centering
		\includegraphics[width=\linewidth]{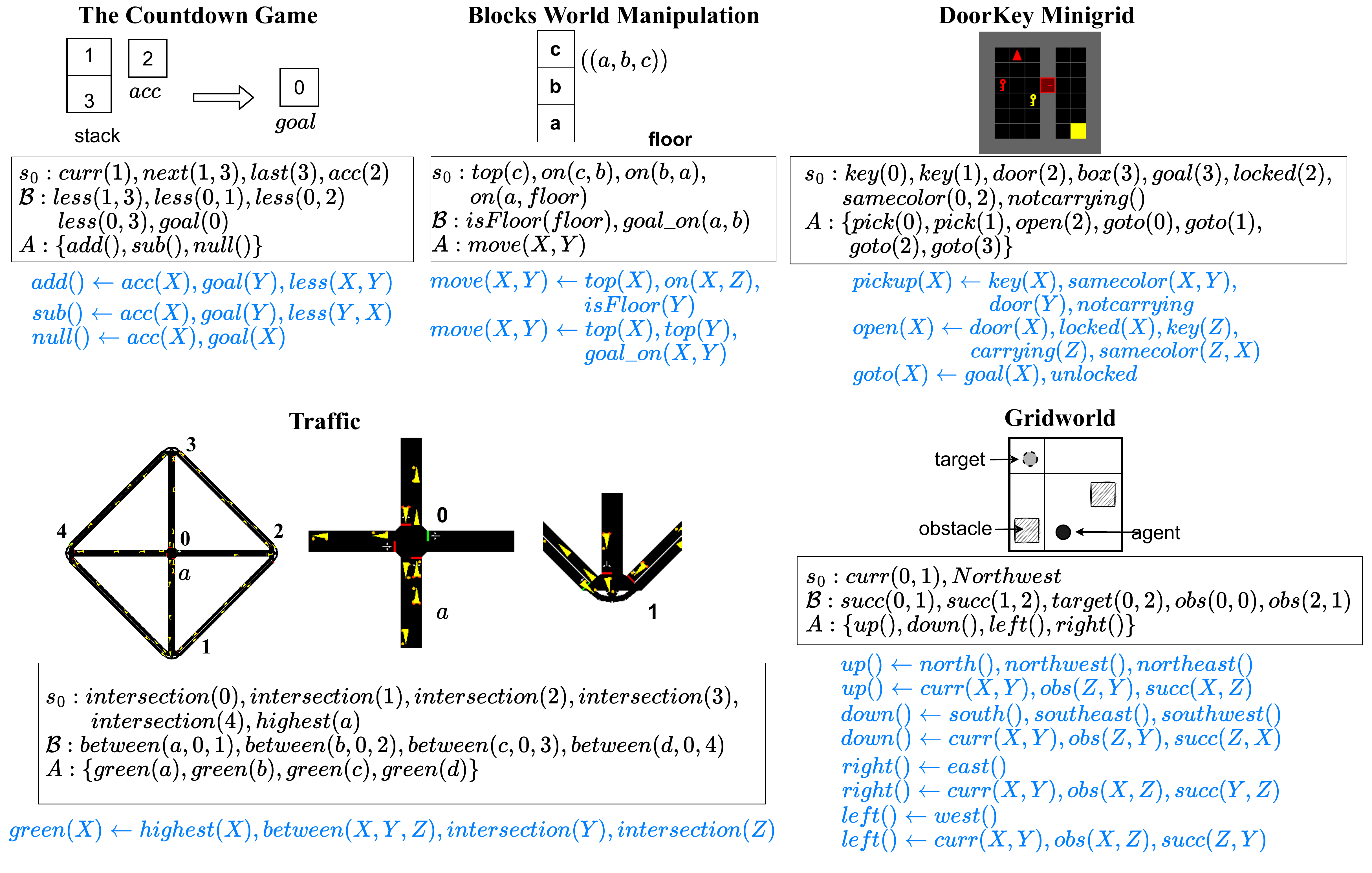}
\caption{Learned rules in all environments. [Row 1]left to right. Countdown Game: select operations (addition, subtraction, null) to make accumulated $=$ target; Blocks World: place a specific block on another; DoorKey: unlock a door with a matching key to reach a goal. [Row 2] Traffic: minimize traffic at grid intersections. The figure shows a 5-agent grid with intersections $0$ and $1$ connected by lane $a$; Gridworld: navigate a grid to reach the goal. Descriptions in Section~\ref{subsection:experimental_setup}.}
    \label{figure:env_desc}
\vspace{-15pt}
\end{figure*}
We introduce \textbf{D}eep \textbf{E}xplainable \textbf{R}elational \textbf{R}einforcement \textbf{L}earning (DERRL: neural DRL + symbolic RRL), a neuro-symbolic RRL approach that combines the strengths of neural (differentiability and representational power) and symbolic methods (generalizability and interpretability) while addressing their respective shortcomings. More specifically, DERRL uses a neural network to search the space of policies represented using First-Order Logic (FOL)-based rules\footnote{DERRL uses a relational representation akin to Quinlan's  FOIL\cite{quinlan_foil} with background knowledge comprising ground facts and non-recursive Datalog-formulated rules.}. Like other ILP methods, our framework  provides interpretable solutions. However, instead of using search-based methods, we leverage the representational capacity of neural networks to generate interpretations of actions (called rules), while entirely bypassing the need to interpret the network itself. To be specific, we propose a parameterized rule generation framework where a neural network learns to generate a set of generalized \textit{discriminative} rules that are representations of the policy. For instance, as shown in the blocks world manipulation game in Figure~\ref{figure:env_desc}, the two rules corresponding to $move$ action are -- $move(X,Y) \leftarrow top(X), on(X,Z), isFloor(Y)$, which triggers an \textit{unstacking} process, and $move(X,Y) \leftarrow top(X), top(Y), goal\_on(X,Y)$, which puts block $a$ on $b$ when both are top blocks in the stacks. Note, that the same rules are applicable for a new goal (say $goal\_on(c,d)$) or when blocks are increased to $10$.

Additionally, we formulate a semantic loss~\cite{semanticloss} to guide the rule learning and restrict the hypothesis space of rules. This loss enforces semantic constraints on rule structure through a differentiable relaxation of semantic refinement~\cite{10.1007/978-3-540-88190-2_1} allowing users to encode background knowledge as axioms to mitigate rule redundancy. For example, in the rule $r \leftarrow less(X,Y), less(Y,Z), less(X,Z)$, due to the transitive relation $less(X,Z) \leftarrow less(X,Y), less(Y,Z)$, the term $less(X,Z)$ is redundant. DERRL enables predefining such knowledge as axioms, penalizing models that violate them. We compare our framework with that of Neural Logic Reinforcement Learning (NLRL)~\cite{pmlr-v97-jiang19a} which uses FOL to represent reinforcement learning policies and is based on Differentiable Inductive Logic Programming ($\partial$ILP) \cite{10.5555/3241691.3241692}. Much like DERRL, NLRL uses policy gradients to train a differentiable ILP module by assigning learnable weights to rules. The authors demonstrate interpretability and generalizability of policies to different problem configurations. We demonstrate DERRL's advantages over NLRL in terms of computational efficiency, policy accuracy, and semantic constraint enforcement.

\noindent\textbf{Contributions.} \textbf{(i)} A neuro-symbolic framework \textbf{DERRL} for learning interpretable RL policies in on-policy, model-free settings, demonstrating their adaptability to environmental changes; \textbf{(ii)} A differentiable relaxation of semantic refinement for guiding rule generation and constraining the hypothesis space.

\section{Related Works}
\label{section:related_works}



\textbf{Integrating Symbolic Planning and RL.} Recent research has sought to merge symbolic planning with deep reinforcement learning (DRL) to improve data efficiency and performance, as seen in works like PEORL~\cite{peorl}, RePReL~\cite{reprel}, and SDRL~\cite{sdrl}. These approaches aim to integrate a high-level planner that suggests sub-goals to be executed by a low-level DRL model, thus relying on predefined environment dynamics, such as high-level action schemas with pre and postconditions. DERRL differs from planning-based methods as it is purely a RL approach, i.e. it does not have access to precise handcrafted action schemas or the reward function, and instead learns suitable control policies through trial-and-error interactions with the environment. Therefore, we only compare DERRL with other RL baselines. \textbf{Explainable RL.} Previous studies on interpretable RL have utilized decision trees for their ease of interpretation. Standard decision trees consist of nested if-then rules, are non-differentiable, and cannot be trained using gradient descent methods. The online nature of RL problems, combined with the non-stationarity introduced by an improving policy, presents additional challenges for decision trees as the agent interacts with the environment. One straightforward but inefficient solution is to re-learn the decision trees from scratch~\cite{JMLR:v6:ernst05a}. More recently, researchers have explored the use of differentiable functions in decision trees~\cite{DBLP:journals/corr/abs-1711-09784}. Differentiable Decision Trees have also been adapted for the RL framework~\cite{pmlr-v108-silva20a,Liu2018TowardID}, although their performance does not match that of deep neural networks. \textbf{Concurrent Works.} Our work on integrating differentiable logic programming into RL is concurrent with efforts such as NLRL~\cite{pmlr-v97-jiang19a} and dNL-ILP~\cite{dNL-ILP}. While dNL-ILP lacks goal generalization, we use the recent NLRL as our baseline. Recent research has also employed Graph Neural Networks~\cite{Kipf:2016tc} to capture relational representations~\cite{garg2019size,garg2020symbolic} with applications to DRL~\cite{gnn_rrl} demonstrating zero-shot generalization to varying problem sizes. DERRL additionally learns interpretable policies. \textbf{Adjusting Language Bias} The possible hypothesis space expands exponentially with input space, necessitating user adjustments to language biases based on domain knowledge. Relational learning systems use declarative bias via semantic refinement~\cite{10.1007/978-3-540-88190-2_1}. Differentiable rule learning methods like $\partial$ILP~\cite{10.5555/3241691.3241692} and NLRL~\cite{pmlr-v97-jiang19a} use rule templates to limit rule body atoms to $2$. However, these methods overlook background knowledge and face redundancies. DERRL mitigates redundancies and shrinks the search space through a differentiable relaxation of semantic refinement.

\section{Preliminaries} 
\label{section:preliminaries}

\subsection{Logic Programming:}
Logic Programming~\cite{10.5555/2214} rules are written as \textbf{clauses} of the form $\alpha \leftarrow \alpha_1, \dots, \alpha_m$ composed of a \textbf{head} atom $\alpha$ and a \textbf{body} $\alpha_1, \dots, \alpha_m$. These clauses are defined using the standard \textit{if-then} rules, wherein, if the body is satisfied, the head is true. Each \textbf{atom} is a tuple $p(v_1, \dots, v_n)$ where $p$ is a n-ary \textbf{predicate} and $v_1, \dots, v_n$ are either \textbf{variables} or \textbf{constants}. A \textbf{ground} atom is one that contains only constants. A predicate can either be \textbf{extensional} when it is defined by a set of ground atoms, or \textbf{target} (intensional) when it is defined by a set of clauses.

An \textbf{alphabet} $\mathcal{L}$ is defined by the tuple $\mathcal{L} := (\mathrm{P}_{\text{tar}}, \mathrm{P}_{\text{ext}}, arity, C, V)$ where, $\mathrm{P}_{\text{tar}}$ is a set of target predicates, $\mathrm{P}_{\text{ext}}$ is a set of extensional predicates, $arity : \mathrm{P}_{\text{ext}} \cup \mathrm{P}_{\text{tar}} \mapsto \mathbb{N}$ is the number of arguments (variables or constants) that the predicate can take, $C$ is a set of constants and $V$ is a set of variables allowed in the clause.
For the blocks world game in Figure~\ref{figure:env_desc}, $\mathrm{P}_{\text{tar}} = \{move/2\}$, $\mathrm{P}_{\text{ext}} = \{top/1, on/2,$ $goal\_on/2, isFloor/1\}, V = \{X,Y,Z\}$, $C=\{a,b,c\}$. 

\subsection{Relational Markov Decision Process:}
We model our problem as a Relational MDP (RMDP) given by the tuple $\mathcal{E}:=(S, \mathcal{B}, A, \delta, r, \gamma)$ which is just like a regular MDP, but with relational states and actions. Here, $S$ is a set of states, where each state is represented as a set of ground atoms consisting of predicates in $\mathrm{P}_{\text{ext}}$ and constants in $C$; $\mathcal{B}$ is the background knowledge also represented in form of ground atoms consisting of predicates and constants, but unlike the state, it remains fixed over the episode; $A$ is a set of actions consisting of ground atoms from predicates in $\mathrm{P}_{\text{tar}}$ and constants in $C$; $\delta: S \times A \mapsto S$ is an \textbf{unknown} transition function; $r: S \times A \mapsto R$ is an \textbf{unknown} real-valued reward function; $\gamma$ is the discount factor.
In the blocks world game (Figure~\ref{figure:env_desc}), for the tuple $((a,b,c))$\footnote{Here, the outer tuple denotes stacks and the inner tuples denote the blocks in the stack. For e.g., $((a,b),(c,d))$ has two stacks: $(a,b)$ is stack 1 and $(c,d)$ is stack 2.}, the initial state $s_0$ and $\mathcal{B}$ are $\{top(c), on(c,b), on(b,a), on(a,floor)\}$ and $\{isFloor(floor), goal\_on(a,b)\}$, respectively. The actions are $move(X, Y)$ where variables $X$ and $Y$ can be substituted with constants in $C$. Although underlying models often use logical transition and reward rules~\cite{kersting2004bellman}, our approach is model-free, so we ignore them here.


\subsection{Problem Statement:}
\textbf{Given}, a tuple $(\mathcal{L}, \mathcal{E})$ where $\mathcal{L}$ is an alphabet, and $\mathcal{E}$ is an RMDP; \\\textbf{Find} an optimal policy $\pi_{\theta}: S \cup \mathcal{B} \mapsto A$ as a set of clauses (also called \textbf{rules}) that maximizes the expected reward $\mathbb{E}_{\tau \sim \pi_{\theta}}\big[R_{\tau}\big]$, where $R_{\tau} := \sum_{k=t+1}^{T-1} \gamma^{k-t-1}R_k$. Here, an episode trajectory is denoted by $\tau$. 

More formally, the rules are selected from the hypotheses space, which is the set of all possible clauses. The head atom of each such rule is an action and the body is the definition of the action. As shown in Figure~\ref{figure:env_desc}, a rule for $move(X,Y)$ in the blocks world environment is given as $move(X,Y) \leftarrow top(X), on(X,Z),$ $isFloor(Y)$ which states that $move(X,Y)$ is triggered when the rule definition (i.e., the body) is satisfied. Thus, if the policy selects the action $move(c,floor)$, one can quickly inspect the body to find out \textbf{how} that action was taken. The rules are \emph{discriminative} (i.e. that help select the correct action by distinguishing it from alternative actions) and together provide an interpretation of the policy. A set of rules is learned for each action. Once trained, the rules for actions do not change and the rule body decides which action should be triggered at each time-step. In what follows, we provide a detailed explanation of rule generation and inference for each time-step of an episode. For simplicity, we expunge the time-step notation (for e.g., state at $t^{th}$ time-step $s_t$ is now $s$).
    

\section{Proposed Approach}
\label{section:proposed_approach}
Consider an alphabet $\mathcal{L}$ where $\mathrm{P}_{\text{tar}}=\{r/0, s/0\}, \mathrm{P}_{\text{ext}}=\{p/1, q/2\}, V= \{X,Y\}, \\
C=\{a,b\}$. The set of all ground atoms $G$ formed from the predicates in $\mathrm{P}_{\text{ext}}$ and constants in $C$ is $\{p(a), p(b), q(a,a), q(a,b), q(b,a), q(b,b)\}$. We represent each ground atom $g_j \in G$ along with its index $j$ in Table~\ref{tab:ground_atoms}.

 Recall, that both state $s$ and background knowledge $\mathcal{B}$ are represented using ground atoms. Given a state $s=\{p(a), q(a,a), q(a,b)\}$ at each time-step, and an empty background knowledge $\mathcal{B}$, we encode it to a state vector $\bm{v}$, such that each element $v_j=1 \; \text{if} \; g_j \in \{s, \mathcal{B}\}$ (i.e. if the current state $s$ or the background knowledge $\mathcal{B}$ contains the ground atom $g_j$), else $0$. Let us now consider the set of all atoms $K$ formed from the predicates in $\mathrm{P}_{\text{ext}}$ and variables in $V$ (instead of the constants in $C$). Table~\ref{tab:all_atoms} lists the atoms $k_j \in K$ and the corresponding $\bm{v}$.


\begin{table}[h]
\vspace{-20pt}
\centering
\caption{Table of all ground atoms $G$ and their indices $j$.}
\small{\begin{tabular}{c c c c c c c | l}
\toprule
     $j$ & 0 & 1 & 2 & 3 & 4 & 5 & \\
     $g_j$ & $\mathbf{p(a)}$ & $p(b)$ & $\mathbf{q(a,a)}$ & $\mathbf{q(a,b)}$ & $q(b,a)$ & $q(b,b)$ & $s=\{p(a), q(a,a), q(a,b)\}$\\
     $v_j$ & \textbf{1} & 0 & \textbf{1} & \textbf{1} & 0 & 0 & $\bm{v} = [1, 0, 1, 1, 0, 0]$\\
\bottomrule
\end{tabular}}
\vspace{-20pt}
\label{tab:ground_atoms}
\end{table}

\begin{table}
\vspace{-20pt}
\centering
\caption{All atoms $K$, their indices $j$, the generated rule vectors ${\bm{b}^r, \bm{b}^s}$, and corresponding probability vectors ${\bm{P}^r, \bm{P}^s}$ for target actions $\mathrm{P}_{\text{tar}}={r/0, s/0}$.}
\small{\begin{tabular}{c c c c c c c l}
\MyToprule(lr){1-8}
     $j$ & $0$ & $1$ & $2$ & $3$ & $4$ & $5$ \\
     $k_j$ & $\textcolor{blue}{p(X)}$ & $\textcolor{red}{p(Y)}$ & $q(X,X)$ & $q(X,Y)$ & $\textcolor{red}{q(Y,X)}$ & $\textcolor{blue}{q(Y,Y)}$ \\
    \midrule
     $b_j^{r}$ & 0 & \textcolor{red}{1} & 0 & 0 & \textcolor{red}{1} & 0 & $r \leftarrow \textcolor{red}{p(Y), q(Y,X)}$\\
     $P_j^{r}$ & 0.1 & \textcolor{red}{0.8} & 0.3 & 0.4 & \textcolor{red}{0.7} & 0.2 & $\bm{w}^r= [\textcolor{red}{0.8, 0.7}]^\top$\\
    \midrule
     $b_j^{s}$ & \textcolor{blue}{1} & 0 & 0 & 0 & 0 & \textcolor{blue}{1} & $s \leftarrow \textcolor{blue}{p(X), q(Y,Y)}$\\
     $P_j^{s}$ & \textcolor{blue}{0.6} & 0.3 & 0.4 & 0.2 & 0.1 & \textcolor{blue}{0.9} & $\bm{w}^s= [\textcolor{blue}{0.6, 0.9}]^\top$\\
\MyBottomrule(lr){1-8}
\end{tabular}}
\vspace{-15pt}
\label{tab:all_atoms}
\end{table}

We represent rules using rule vectors. As shown in Table~\ref{tab:all_atoms}, the rule vector for each action $i \in A$ is given as $\bm{b}^i \in \{0,1\}^m$, where $m=\mid K \mid$ (i.e. the cardinality of the set of all atoms formed from the predicates in $\mathrm{P}_{\text{ext}}$ and variables in $V$).  Here, $b_j^i=1$, if the $j^{th}$ atom is in the body of the $i^{th}$ rule. From Table~\ref{tab:all_atoms}, $\bm{b}^{r} = [0, \textcolor{red}{1}, 0, 0, \textcolor{red}{1}, 0]^\top$ corresponds to the rule $r \leftarrow \textcolor{red}{p(Y), q(Y,X)}$.

We impose the Object Identity (OI) assumption~\cite{objectIdentity} which states that during grounding and unification, distinct variables must be mapped to distinct constants. For instance, ground rules for $r \leftarrow p(Y), q(Y,X)$ are $r \leftarrow p(b), q(b,a)$ and $r \leftarrow p(a), q(a,b)$ under substitutions $\phi_0 = \{a/X, b/Y\}$ and $\phi_1=\{b/X,a/Y\}$, respectively, but a substitution $\phi_2 = \{a/X, a/Y\}$ is not allowed. Without loss of generality, one can model nullary predicates and negated atoms, by simply including additional dimensions (corresponding to atoms in $K$) in the vector $\bm{b}^i$.

   
    

The DERRL framework learns a rule vector $\bm{b}^i$ for each action $i \in A$ by associating it with a trainable weight vector $\bm{w}^i$. Each element $w_j^i \in \bm{w}^i$ indicates the membership of the corresponding atom in the rule definition (i.e., if the weight of the atom is high, it is more likely to belong to the rule definition). Given the state vector $\bm{v}$, action probabilities $\pi_{\theta}(i \mid s, \mathcal{B})$ are calculated by performing a fuzzy conjunction on the rules~(Section~\ref{subsubsection:fuzzy_conjunction_operators}). The whole framework is trained end-to-end using the REINFORCE algorithm~\cite{Williams:1992:SimpleStatisticalGradientFollowingAlgorithmsForConnectionistReinforcementLearning}, with the loss function given as $J(\pi_{\theta}) = -\mathbb{E}_{\tau \sim \pi_{\theta}} \big[R_{\tau}\big]$. Here, $R_{\tau}$ is the discounted sum of rewards over trajectory $\tau$, and $\theta$ is the set of trainable parameters.

Algorithm~\ref{algorithm:rule_generation} summarizes the DERRL framework. The two main components of DERRL are \textbf{(i)}  Rule Generator (Section~\ref{subsection:rule_generation}), which at every time-step $t$ and for each action $i \in A$, generates a rule vector $\bm{b}^i$, and a weight vector $\bm{w}^i$; \textbf{(ii)} Forward chaining Inference (Section~\ref{subsection:inference}) that takes the generated rules vectors for all actions $\{\bm{b}^i\}_{i=1}^{\mid  A \mid }$, the corresponding weight vectors $\{\bm{w}^i\}_{i=1}^{\mid  A \mid }$, and the state valuation vector $\bm{v}$ for the $t^{th}$ time-step, and returns the action probabilities $\pi_{\theta}(. \mid  s, \mathcal{B})$. Note, that the rule generator parameters $\theta$ are trained by calculating the gradients of the loss function with respect to weight vectors~$\bm{w^i}$. 



\subsection{Rule Generation}
\label{subsection:rule_generation}

The rule $\mathcal{R}_{\theta}: i \mapsto \bm{b}^i, \bm{w}^i$ uses a parameterized network $\mathcal{R}_{\theta}$ to map each action (index) $i$ to a rule vector $\bm{b}^i$ and weight vector $\bm{w}^i$. Here, $b_j^i=1$ indicates the $j^{th}$ atom is in the rule body. The rule generator outputs a probability vector $\bm{P}^i$ where $P_j^i$ represents the probability of the $j^{th}$ atom in $K$ belonging to the rule body. We use the Gumbel-max trick~\cite{JangEtAl:2017:CategoricalReparameterizationWithGumbelSoftmax} on $\bm{P}^i$ to sample the binary vector $\bm{b}^i$. 
\begin{equation*}
    b_j^i = \argmax(\log(P_j^i) + u_0, \log(1 - P_j^i) + u_1) \; \text{where} \; u \sim \Gumbel(0,1)
\end{equation*}
Here, $\Gumbel(0,1)$ is the standard Gumbel distribution given by the probability density function $f(x) = e^{-(x + e^{-x})}$. During evaluation, we use $\argmax(.)$ operation without sampling. From $\bm{P}^i$, we also obtain the weight vector $\bm{w}^i \in \mathbb{R}^{\| \bm{b}^i \|_1}$ comprising the probabilities of only those atoms which have $b_j^i=1$. From Table~\ref{tab:all_atoms}, the generated rule vector $\bm{b}^r = [0, \textcolor{red}{1}, 0, 0, \textcolor{red}{1}, 0]^\top$, the probability vector $\bm{P}^r = [0.1, \textcolor{red}{0.8}, 0.3, 0.4, \textcolor{red}{0.7}, 0.2]^\top$, and the corresponding weight vector $\bm{w}^r= [\textcolor{red}{0.8}, \textcolor{red}{0.7}]^\top$. 


\subsection{Inference}
\label{subsection:inference}

\begin{center}
\small{\begin{tabular}{c c c c c c c}
\toprule
    $i$ & $\mathbf{X}^i$ & $\mathbf{Y}^i$ & $\bm{w}^i$ & $\bm{z}^i$& $\mathcal{F}^i$ & $\pi_{\theta}(i \mid s, \mathcal{B})$\\
\toprule
     $r \leftarrow \textcolor{red}{p(Y), q(Y,X)}$ & $\begin{bmatrix}
                    0,3\\
                    1,4
                    \end{bmatrix}$  & $\begin{bmatrix}
                                            1,1\\
                                            0,0
                                            \end{bmatrix}$  &  $\begin{bmatrix}
                                                                0.8\\
                                                                0.7
                                                                \end{bmatrix}$ 
                                            & $\begin{bmatrix}
                                                    0.5\\
                                                    0
                                                    \end{bmatrix}$ & $0.5$ & $0.62$\\
    \midrule
    $s \leftarrow \textcolor{blue}{p(X), q(Y,Y)}$ & $\begin{bmatrix}
                    0,5\\
                    1,2
                    \end{bmatrix}$ & $\begin{bmatrix}
                                            1,0\\
                                            0,1
                                            \end{bmatrix}$  &  $\begin{bmatrix}
                                                                0.6\\
                                                                0.9
                                                                \end{bmatrix}$
                                           & $\begin{bmatrix}
                                                    0\\
                                                    0
                                                    \end{bmatrix}$ & $0$ & $0.38$\\
    \bottomrule
\end{tabular}}
\end{center}

\noindent Consider the following rules generated by the rule generator. Each rule is passed through a substitution $\phi: V \mapsto C$ to produce ground rules. 
Let, $\mathbf{X}^i \in \mathbb{Z}_{\geq 0}^{N(\phi) \times \|\bm{b^i}\|_1}$ be the matrix representation of the ground rules. Here, $\mathbb{Z}_{\geq 0}$ and $N(\phi)$ are the set of non-negative integers and the number of possible substitutions, respectively. Each row in $\mathbf{X}^i$ is a vector of ground atom indices that belong to the ground rules. Using a substitution $\phi=\{b/X,a/Y\}$, we obtain the ground rule $r \leftarrow p(a), q(a,b)$. From Table~\ref{tab:ground_atoms}, this rule definition can be written as the vector $[0,3]$ (i.e., indices of $p(a)$ and $q(a,b)$ are 0, 3, respectively). Similarly, the substitution $\phi=\{a/X,b/Y\}$ gives us  $r \leftarrow p(b), q(b,a)$, and the vector $[1,4]$.

Next, the $\Value(.)$ operation takes each element $\mathbf{X}^i_j$ and returns its state value $v_j$. From Table~\ref{tab:ground_atoms}, $    \mathbf{Y}^r = \Value(\mathbf{X}^r)= \Value(\begin{bmatrix}0,3\\1,4\end{bmatrix}) = \begin{bmatrix}v_0,v_3\\v_1,v_4\end{bmatrix} = \begin{bmatrix}1,1\\0,0\end{bmatrix}$. The row vectors of $\mathbf{Y}^i = [\bm{y}_1^i, \dots, \bm{y}_{N(\phi)}^i] \in \mathbb{R}^{N(\phi) \times \|\bm{b^i}\|_1}$ can be regarded as truth values of the grounded rule (i.e., for each substitution $\phi$), based on $\{s,\mathcal{B}\}$. If the rule definition is not satisfied, the corresponding row vectors will have sparse entries. To ensure differentiability, we use fuzzy norms for our rules.

\subsubsection{Fuzzy Conjunction Operators.}
\label{subsubsection:fuzzy_conjunction_operators}
Fuzzy norms integrate logic reasoning with deep learning by approximating the truth values of predicates~\cite{10.5555/3241691.3241692,Marra2019TNormsDL}. Fuzzy conjunction operators $* : [0,1]^{\|\bm{b^i}\|_1} \mapsto [0,1]$ can be of various types like Godel t-norm and Product t-norm (refer Section~\ref{appendix:fuzzy_conjunction_operators}). We use Lukasiewicz t-norm ($\top_{\text{Luk}}(a,b) := \Max\{0, a+b-1\}$) to compute the action values for each rule\footnote{More generally, given a vector $\bm{y} \in [0, 1]^n$, Lukasiewicz t-norm $\top_{\text{Luk}}\bm{y} := \Max(0, \langle \bm{y}, \mathbb{1} \rangle - n + 1)$. For the proof, refer Section~\ref{appendix:fuzzy_conjunction_operators}.}. To encourage the rule generator to generate more precise rules with higher probability, we calculate a valuation vector by weighing each row $\bm{y}_k^i \in \mathbb{R}^{\|\bm{b^i}\|_1}$ with the weight vector $\bm{w}^i \in \mathbb{R}^{\|\bm{b^i}\|_1}$, and using the Lukasiewicz operator as $z_k^i = \Max(0, \langle \bm{y}_k^i, \bm{w}^i \rangle - \mid \bm{w}^i \mid + 1)$. Intuitively, the inner product $\langle \bm{y}_k^i, \bm{w}^i \rangle$ is a weighted sum over all atoms in the rule body that are true in $s \cup \mathcal{B}$. This is akin to performing $(a+b)$ in t-norm operator. For $\bm{y}_0^r = [1, 1]^{\top}, \bm{w}^r = [0.8, 0.7]$:
\begin{equation*}
    z_0^r = \Max(0, \langle \begin{bmatrix} 1\\ 1 \end{bmatrix}, \begin{bmatrix} 0.8\\ 0.7 \end{bmatrix}\rangle - 1) = 0.5
\end{equation*}
With multiple substitutions (or groundings) for a generated rule, we find the maximum valuation as $\mathcal{F}^i = \Max(\bm{z}^i)$. The final action probability is calculated as $\pi_{\theta}(i \mid s, \mathcal{B}) = \softmax(\mathcal{F}^i)$. Note, that if the generated rule is not satisfied for any substitution (i.e., has a sparse row vector in the matrix $\mathbf{Y}^i$), the valuation of the generated rule is lower (for e.g., from the above table $\mathcal{F}^s < \mathcal{F}^r$).


\subsubsection{Multiple rules for a single action.}
\label{subsubsection:multiple_rules_for_single_action}
We generalize DERRL to learn policies with multiple rules for each action, allowing it to switch between rules based on input (for e.g. in the blocks world game, the "move" action uses two rules executed at different steps depending on the goal blocks' position). We allow multiple rule networks per action, adjusting the final computation step to determine action probabilities based on the best-satisfied rule. Given $\mathcal{F}_1^i$ and $\mathcal{F}_2^i$ for two different rules for the same action, we first compute $\tilde{\mathcal{F}^i} = \Max(\mathcal{F}_1^i, \mathcal{F}_2^i)$ to determine which rule is more appropriate at a given time-step. Consider two different rules generated for action $r$ with arbitrary $\mathbf{Y}^i$:

\begin{center}
\small{\begin{tabular}{c c c c c}
\toprule
    $i$ & $\mathbf{Y}^i$ & $\bm{w}^i$ & $\bm{z}^i$& $\mathcal{F}^i$\\
\toprule
     $r_1 \leftarrow p(Y), q(Y,X)$ & $\begin{bmatrix}
                    1,1\\
                    0,0
                    \end{bmatrix}$  &  $\begin{bmatrix}
                                        0.8\\
                                        0.7
                                        \end{bmatrix}$ 
                    & $\begin{bmatrix}
                            0.5\\
                            0
                            \end{bmatrix}$ & $0.5$\\
    \midrule
    $r_2 \leftarrow p(X), p(Y), q(X,X)$ & $\begin{bmatrix}
                    1,0,1\\
                    0,1,1
                    \end{bmatrix}$  &  $\begin{bmatrix}
                                        0.6\\
                                        0.8\\
                                        0.8
                                        \end{bmatrix}$
                   & $\begin{bmatrix}
                            0\\
                            0\\
                            0
                            \end{bmatrix}$ & $0$\\
    \bottomrule
\end{tabular}}
\end{center}

Here, $\tilde{\mathcal{F}^r} = \Max(0.5, 0) = 0.5$ is the valuation for rule $r$. Intuitively, depending on the current state $s$ and background knowledge $\mathcal{B}$, one of the rules will be more appropriate (i.e., lower sparsity in rows of $\mathbf{Y}^i$) than the others, prompting the policy to switch to that rule for decision making\footnote{This assumes a specified upper bound on the number of rules for each action, similar to selecting the number of clusters in a clustering algorithm~\cite{1427769}.}.


\begin{algorithm}[t]
\caption{Deep Explainable Relational Reinforcement Learning (DERRL)}\label{algorithm:rule_generation}
\textbf{Input:} Alphabet: $\mathcal{L}$, RMDP: $\mathcal{E}$\\
 \textbf{Output:} (set of) rules that encode the policy $\pi_{\theta}$

Initialize  rule generator parameters $\theta$\;

\begin{algorithmic}
\For{each episode}
    \For{$t=0$ to $T-1$}
    
    \State $\bm{v} = \encode(s, \mathcal{B})$\; \Comment{state vector}
    
        \For{each action $i$}
            \State $\bm{b}^i, \bm{w}^i \sim \mathcal{R}_{\theta}(i)$ \Comment{Rule Generation (Section~\ref{subsection:rule_generation})}
        \EndFor
        
        \State $\pi_{\theta}(. \mid s, \mathcal{B})$ = $Inference(\bm{v}, \{\bm{b}^{i}\}_{i=1}^{\mid A \mid}, \{\bm{w}^i\}_{i=1}^{\mid A \mid})$ \Comment{(Section~\ref{subsection:inference})}
        \State $a \sim \pi_{\theta}{(. \mid s, \mathcal{B})}$
        \State $s' \leftarrow \delta(s, a)$; \;\; $R_t \leftarrow r(s, a)$
        
    \EndFor
    
    \State $R_{\tau} \leftarrow \sum_{t+1}^{T-1} \gamma^{k-t-1}R_k $
    \State $\theta \leftarrow \theta - \eta\nabla_{\theta} \mathbb{E}_{\tau \sim \pi_{\theta}} [R_{\tau}]$
\EndFor
\end{algorithmic}
\end{algorithm}
\subsection{Semantic Constraints}
\label{subsection:semantic_constraints}
The set of possible rules to consider grows exponentially with the number of predicates and their arity. While traditional relational learning systems have used declarative bias in form of semantic refinement~\cite{10.1007/978-3-540-88190-2_1}, prior works in differentiable rule learning~\cite{10.5555/3241691.3241692,pmlr-v97-jiang19a} employ rule templates to restrict the hypothesis space (e.g. rules of size $2$). However, these methods frequently encounter redundancies. For example, rules $r \leftarrow less(X,Y), less(Y,Z), less(X,Z)$ and $s \leftarrow equal(X,Y), equal(Y,X)$ exhibit transitive and symmetric relations, respectively, making some atoms redundant. A rule $r$ is redundant w.r.t. a constraint $h \leftarrow b_1, ... , b_n$ if the rule  $false \leftarrow h, b_1, ... , b_n$ subsumes the rule $r$. To avoid redundancies, generated rule vectors with $b^i_j=1$ for both atoms $equal(X,Y)$ and $equal(Y,X)$ should be penalized. To this end, we propose a differentiable relaxation of semantic refinement by applying a supervised loss on probability vectors $\{\bm{P}^i\}_{i=1}^{\mid A \mid}$. We declare semantic constraints $\mathcal{S}_c$ as axioms which can either be a relation (symmetric or transitive), or some background fact (like $false \leftarrow on(X,Y), on(Y,X)$). Then we calculate the semantic loss as $\mathcal{L}_{sem} = \sum_{x \in 
\mathcal{S}_c} \sum_{i \in A} \prod_{j \in x} P^i_j$. 

Here, the outer summation is over each semantic constraint $x \in \mathcal{S}_c$, and the inner summation is over each generated rule $i \in A$. The product is over the probability of each atom (with index $j$) in the body of axiom $x$. For instance, given a single axiom $false \leftarrow p(Y), q(Y,X)$, for the generated rule $r \leftarrow p(Y), q(Y,X)$, from Table~\ref{tab:all_atoms}, the loss is $\mathcal{L}_{sem} = P^i_1 \times P^i_4 = 0.56$. Here, the loss is high because according to the given constraint, $p(Y)$ and $q(Y,X)$ should not appear together in the body of the rule. Intuitively, the loss is highest if the membership probabilities of both atoms are high warranting a penalization. $\mathcal{L}_{sem}$ is summed over the entire episode and the final loss is given as $\bar{J}(\pi_{\theta}) = J(\pi_{\theta}) + \lambda_{sem} \mathcal{L}_{sem}$. Here, $\lambda_{sem}$ is a regularization term. See Appendix~\ref{appendix:semantic_constraints} for constraints in all environments.




\section{Experiments}
\label{section:experiments}

Through our experiments, we aim to answer the following questions: (1) Can the proposed approach learn interpretable policies while performing on par with neural-based baselines? (Section~\ref{subsection:interpretation_of_policies}); (2) Are the learned rules agnostic to modifications in the environment? (Section~\ref{subsection:generalization_performance}); (3) How efficient and scalable is the proposed approach compared to the current state-of-the-art NLRL? (Section~\ref{subsection:comparison_with_nlrl})

\subsection{Experimental Setup}
\label{subsection:experimental_setup}

\textbf{The Countdown Game.} The agent manipulates a stack of numbers and an initial accumulated value $acc(X)$ to match a target number $goal(X)$ by applying operations like addition ($add$), subtraction ($sub$), or no operation ($null$). The stack comprises of the top number ($curr(X)$), number below it ($next(X,Y)$), and bottom-most number ($last(X)$). From Figure~\ref{figure:env_desc}, the state $s_t$ includes the stack, accumulated number, and goal number. Operations are performed between the accumulated value and the top number of the stack\footnote{$add$: acc $\pluseq$ top, $sub$: acc $\subeq$ top, $null$: acc}. The background knowledge $\mathcal{B}$ comprises the target number\footnote{$goal(X)$ is provided as background since it does not change during the episode.}, and atoms of the form $less(X,Y)$ which denote that number $X$ is less than $Y$. A reward of $r=1$ is given when the target and accumulated values match at the end of the episode, otherwise $r=-\frac{|goal-acc|}{N_1}$ where $N_1$ is a normalizing constant. An initial range of numbers $[-4, 6]$ and stack of length$=2$ is used for training. The learned models are tested for generalization on the following tasks (i) dynamic stack lengths of $\{3, 4, 5\}$; (ii) held-out target unseen during training; (iii) held-out initial stack sequences. We also train a stochastic game version with $10 \%$~probability of altering an action to null. 

\noindent\textbf{Blocks World Manipulation.} Given an initial configuration of blocks, the goal is to put a specified block atop another specified block (Figure~\ref{figure:env_desc}). Stacks are represented using predicates: $top(X)$ means that block $X$ is the top block, $on(X,Y)$ means that block $X$ is on top of block $Y$. The actions are $move(X,Y)$ with $X=\{a,b,c\}$ and $Y=\{a,b,c,floor\}$. A reward $r=1$ is provided if the task is achieved. To enforce optimal planning, we impose a penalty of $r=-0.02$ for every action. Training includes a fixed number of blocks $=3$ and a fixed goal -- to stack block $a$ on block $b$ ($goal\_on(a,b)$). We train it with initial configurations: $((a,b,c)); ((c,a,b)); ((a,c),(b)); ((b,c), (a))$. Here, each tuple is a stack. For generalization, we use variations like (i) held-out configuration unseen during training like $((a,b),(c)); ((b,c,a)); ((b,a,c))$; (ii) dynamic (number of) blocks $\{4, 5\}$; (iii) dynamic (number of) stacks $\{2,3,4\}$; (iv) unseen goals like $goal\_on(b,a)$ and $goal\_on(a,c)$.

\noindent\textbf{Gridworld.} The agent navigates a grid with obstacles to reach the goal (Figure~\ref{figure:env_desc}). The agent can move vertically ($up/down$) or sideways ($left/right$). The state information consists of the current position of the agent $curr(X,Y)$ where $X$ and $Y$ are the coordinates, and the compass direction of the target (\textit{North, South, East, West, Northeast, Northwest, Southeast, Southwest}). The background information consists of target coordinates ($target(X,Y)$), obstacle coordinates ($obs(X,Y)$), and successor information $succ(X, Y)$ where $Y=X+1$. The action space is $\{up, down, left, right\}$. The agent receives a reward of $r=1$ for reaching the target, otherwise $r=-\frac{\| position_{goal} - position_{agent}\|_2}{N_2}$. Here $N_2$ is a normalizing constant. During training, a fixed size grid of $3\times3$ and $5\times5$ is used with the number of obstacles being fixed $=2$. For generalization, we use the following variations: (i) dynamic (number of) obstacles $\{3, 4\}$; (ii) held-out (agent-goal) configurations. Unlike graph search algorithms like $\mathrm{A}^{\ast}$ that assume access to the dynamics model, DERRL learns actions through exploration.

\noindent\textbf{Traffic.} We used the Simulation of Urban MObility (SUMO) traffic simulator~\cite{sumo} to simulate traffic flow, where intersections (3-way and 4-way) function as agents denoted by $intersection(Y)$, and are connected by a network of 2-way lanes represented as $between(X,Y,Z)$, indicating a connection between intersections $Y$ and $Z$ by lane $X$.  The goal is to minimize the traffic at the intersections, hence reward is the negative queue length at each intersection. Each agent is provided with the lane that has the highest traffic, labeled as $highest(X)$ for lane $X$, and is responsible for controlling the traffic lights for that lane, enabling them to turn the lights green ($green(X)$) for a specific lane $X$. Therefore, 3 and 4-way intersections have an action space of size $3, 4$ respectively. Although only two models (one for all 3-way intersections and another for all 4-way intersections) suffice, we train each intersection independently to demonstrate the scalability of DERRL to multi-agent setups, and also for future developments in cooperative multi-agent setups~\cite{zhang2018fully,gupta2020networked}. We train on a grid comprising 5 agents and transfer the learned rules to an 8-agent grid. We show the mean rewards of all agents in Table~\ref{table:gen_results} -- note, that the best possible reward $\approx 0$.

\noindent\textbf{DoorKey Minigrid.} The agent task is to unlock a locked door ($locked(Y), \\door(Y)$) and reach a goal ($goal(Z)$). Various colored keys ($key(X)$) are scattered throughout the room, and the agent must select the key that matches the color of the door ($samecolor(X,Y)$) to unlock it. We use high-level actions. The agent is only allowed to carry one key at a time and can navigate to and pick up a key $X$ using the $pick(X)$ action if it is not carrying any keys ($notcarrying$), otherwise, it drops the key before picking the new one. The $open(X)$ action enables it to unlock a door $X$ if it carries the key that matches the door's color. The $goto(X)$ action enables the agent to navigate to a specific object $X$. The reward $=1$ for successfully reaching the goal, else $0$. The learned model is tested for generalization with additional doors and keys of varying colors.

We evaluate our DERRL against Neural Logic Reinforcement Learning \\(\textbf{NLRL}) baseline. We also compare with model-free DRL approaches with variations in the deep learning module like (i) Graph Convolution Network (\textbf{GCN})~\cite{Kipf:2016tc} that perform well at relational learning~\cite{ijcai2020-679}, and are invariant to the number of nodes in the graph; (ii) Multilayer Perceptron (\textbf{MLP}). Finally, we compare with a Random (\textbf{Random}) baseline where the weights of the MLP are randomized to set a lower limit on the performance. We use a single-layer neural network for our rule generator (2$m$ parameters). For the GCN and MLP baselines, we experimented with 2-layer networks (O($m^3$) parameters).
\begin{table*}[t]
    \centering
    \caption{Generalization Scores (average rewards over 50 episodes across 3 runs) compare DERRL to other baselines. DERRL outperforms baselines, including the state-of-the-art NLRL. In Traffic, the mean reward for all agents is reported.}
    \small{\begin{tabular}
    {>{\centering\arraybackslash}m{2.0cm} >{\centering\arraybackslash}m{3.0cm}
    >{\centering\arraybackslash}m{1.3cm}
    >{\centering\arraybackslash}m{1.3cm}
    >{\centering\arraybackslash}m{1.3cm}
    >{\centering\arraybackslash}m{1.3cm}
    >{\centering\arraybackslash}m{1.3cm}}
        \toprule 
        \textbf{Setup} & \textbf{} & \textbf{DERRL} & \textbf{NLRL} & \textbf{GCN} & \textbf{MLP} & \textbf{Random}\\ 
        \midrule
        Countdown Game & 
        \begin{tabular}{>{\centering\arraybackslash}m{3.0cm}>{\centering\arraybackslash}m{1.3cm}>{\centering\arraybackslash}m{1.3cm}>{\centering\arraybackslash}m{1.3cm}>{\centering\arraybackslash}m{1.3cm}>{\centering\arraybackslash}m{1.3cm}} training & 0.98 & 0.95 & 0.98 & \textbf{1.00} & 0.30 \\ \hline dynamic stack & \textbf{0.98} & 0.95 & 0.95 & 0.54 & 0.38 \\ \hline held-out target & \textbf{0.98} & 0.85 & 0.95 & 0.35 & 0.33 \\ \hline held-out initial & 0.98 & 0.55 & \textbf{1.00} & 0.35 & 0.18 
        \end{tabular} \\
        \midrule
        Countdown Game(stochastic) & \begin{tabular}{>{\centering\arraybackslash}m{3.0cm}>{\centering\arraybackslash}m{1.3cm}>{\centering\arraybackslash}m{1.3cm}>{\centering\arraybackslash}m{1.3cm}>{\centering\arraybackslash}m{1.3cm}>{\centering\arraybackslash}m{1.3cm}} training & 0.98 & 0.95 & 0.98 & \textbf{1.00} & 0.15
        \end{tabular} \\
        \midrule
        Blocks World Manipulation & \begin{tabular}{>{\centering\arraybackslash}m{3.0cm}>{\centering\arraybackslash}m{1.3cm}>{\centering\arraybackslash}m{1.3cm}>{\centering\arraybackslash}m{1.3cm}>{\centering\arraybackslash}m{1.3cm}>{\centering\arraybackslash}m{1.3cm}} training & \textbf{0.97} & \textbf{0.97} & \textbf{0.97} & \textbf{0.97} & -0.18 \\ \hline held-out config. & \textbf{0.97} & 0.70 & 0.55 & 0.45 & -0.18 \\ \hline dynamic blocks & \textbf{0.92} & 0.51 & -0.20 & -0.21 & -0.22\\ \hline dynamic stacks & \textbf{0.96} & 0.90 & 0.90 & 0.85 & -0.18\\ \hline unseen goal & \textbf{0.96} & 0.45 & -0.18 & -0.18 & -0.18
        \end{tabular} \\
        \midrule
         DoorKey Minigrid & \begin{tabular}{>{\centering\arraybackslash}m{3.0cm}>{\centering\arraybackslash}m{1.3cm}>{\centering\arraybackslash}m{1.3cm}>{\centering\arraybackslash}m{1.3cm}>{\centering\arraybackslash}m{1.3cm}>{\centering\arraybackslash}m{1.3cm}} training & 0.80 & 0.45 & 0.75 & \textbf{0.90} & 0.10 \\ \hline dynamic keys/doors & \textbf{0.78} & 0.25 & 0.35 & 0.20 & 0.05
        \end{tabular} \\
        \midrule
        Traffic & \begin{tabular}
        {>{\centering\arraybackslash}m{3.0cm}>{\centering\arraybackslash}m{1.3cm}>{\centering\arraybackslash}m{1.3cm}>{\centering\arraybackslash}m{1.3cm}>{\centering\arraybackslash}m{1.3cm}>{\centering\arraybackslash}m{1.3cm}} training (5-agents) & \textbf{-0.76} & -0.91 & -0.90 & -0.95 & -1.54 \\ \hline 8-agents &  \textbf{-1.02} & -1.28 & -1.45 & -1.75 & -2.17
        \end{tabular} \\
        \midrule
        Gridworld Game & \begin{tabular}
        {>{\centering\arraybackslash}m{3.0cm}>{\centering\arraybackslash}m{1.3cm}>{\centering\arraybackslash}m{1.3cm}>{\centering\arraybackslash}m{1.3cm}>{\centering\arraybackslash}m{1.3cm}>{\centering\arraybackslash}m{1.3cm}} training & 0.75 & 0.72 & 0.70 & \textbf{0.81} & 0.03\\ \hline dynamic obstacles & \textbf{0.70} & 0.55 & 0.46 & 0.51 & -0.15 \\ \hline held-out config. & \textbf{0.81} & 0.70 & 0.17 & -0.61 & -0.70
        \end{tabular} \\
        \bottomrule
    \end{tabular}}
    \label{table:gen_results}
    \vspace{-10pt}
  \end{table*}

\section{Results}
\label{section:results}

\subsection{Interpretation of policies (Q1)}
\label{subsection:interpretation_of_policies}

In this section, we provide interpretations of the learned rules in each training environment, as shown in Figure~\ref{figure:env_desc}. \textbf{The Countdown Game.}~The policy selects $add$ action when $acc(X) < goal(Y)$, $sub$ action when $acc(X) > goal(Y)$, and null action when both are equal. \textbf{Blocks World Manipulation.}~The policy learns two rules for $move(X,Y)$. Given a goal to put block $a$ atop block $b$, the first rule is applicable when at least one of the blocks is not the top block. Hence, the policy learns to unstack the blocks -- the top block $X$ of the stack is moved to the floor $Y$. The second rule is applied when both $a$ and $b$ are at the top.  \textbf{Traffic.}~The general rule for each intersection is $green(X) \leftarrow highest(X), between(X,Y,Z), intersection(Y), intersection(Z)$. Intuitively, the lights corresponding to the lane with the highest traffic $X$, connecting intersections $Y$ and $Z$, are turned green. \textbf{DoorKey Minigrid.}~The learned rule for action $pickup(X)$ tells the agent to pickup the key $X$ that matches the color of the door $Y$, provided that it is not carrying any other items ($notcarrying$). Similarly, for $open(X)$, the learned rule states that the agent can unlock a locked door $X$ using the key $Y$ only if the colors of the door and the key match. The $goto(X)$ action directs the agent to navigate to the goal object $X$ when the door is $unlocked$. \textbf{Gridworld Game.}~In this setting, the policy learns two rules for each action. The first is used for navigation to the target, such as moving $up$ if the target is to the north (or northeast and northwest) of the grid. The second helps navigate around obstacles, e.g. move $up$ if the obstacle (given by $(Z,Y)$) is to the immediate right of the agent (given by $(X,Y)$). However, the policy may not follow the shortest path\footnote{When the target is to the southeast, and the agent encounters a target to its right, it will  travel north ($up$) rather than south ($down$).} or have consistent traversal strategies, resulting in varied rules for different instances without performance loss.

DERRL also bears similarities with Program Synthesis~\cite{10.1145/321992.322002,4310035,PGL-010}, which involves finding a program that meets user specifications and constraints. Learned policies can be rewritten as programs (see Appendix~\ref{appendix:analogies_with_program_synthesis_cont}). For example, a program to solve the countdown game involves operating on current and accumulated values (using $add, sub, null$ operations) until accumulated value $=$ goal value.

\begin{figure*}[t]
    \centering
    \includegraphics[width=0.9\linewidth]{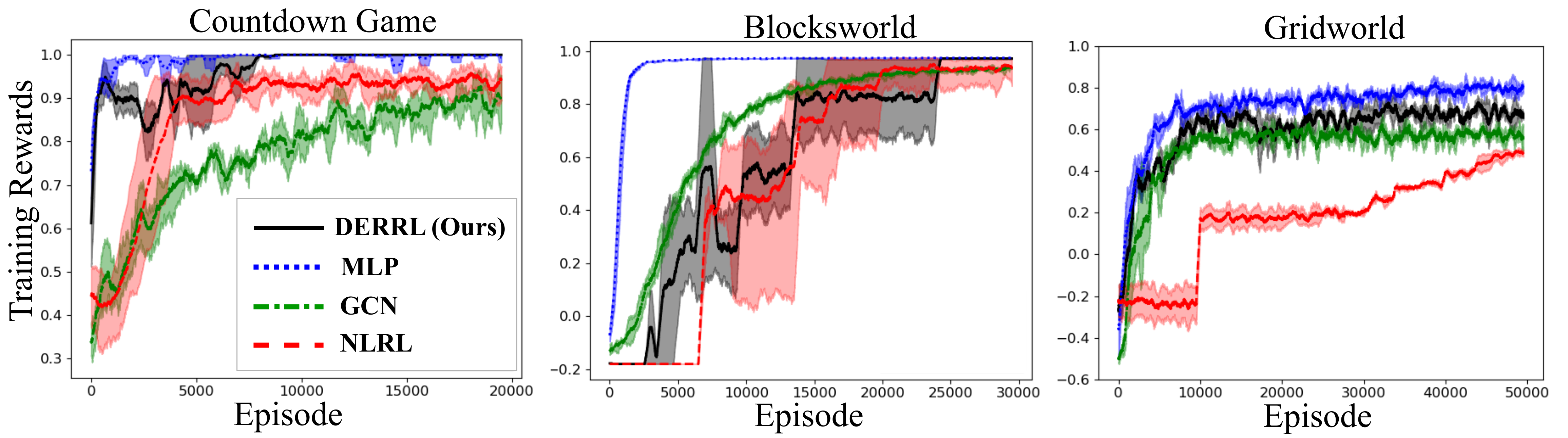}
            \vspace{-1em}
    \caption{[Best viewed in color]~Comparison of training rewards at convergence for different baselines plotted by averaging the rewards over 3 independent runs. [Left to right]~Countdown Game, Blocks World, and Gridworld.}
	\label{figure:convergence_plots}
 \vspace{-10pt}
\end{figure*}

\subsection{Generalization performance (Q2)}
\label{subsection:generalization_performance}
From Table~\ref{table:gen_results}, we observe that DERRL learns general rules, generalizing to environment modifications, and outperforms baselines in generalization tasks. Unlike symbolic planning, DERRL's performance remains unaffected by noisy training, such as stochastic Countdown. Secondly, although GCNs perform well in relational learning, their generalization is marginally better than MLP, potentially failing to capture task agnostic relational patterns. However, GCN surpasses MLP and NLRL in the countdown game. Lastly, DERRL's training performance is comparable to MLP, but it outperforms MLP in generalization tasks, where MLP is similar to the Random baseline. Additionally, DERRL's convergence speed is on par with MLP, as shown in Figure~\ref{figure:convergence_plots}.


\subsection{Comparison with NLRL (Q3)}
\label{subsection:comparison_with_nlrl}
\begin{figure}[t]
	\centering
        \includegraphics[width=0.9\linewidth]{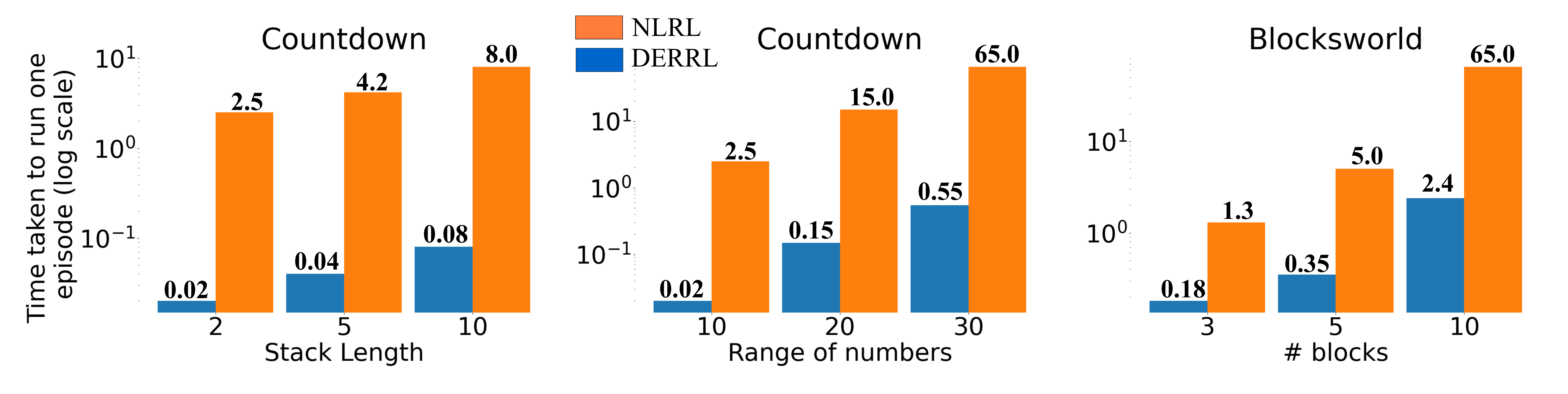}
        \vspace{-1em}
		\caption{Run-time comparison per episode between DERRL and NLRL as problems scale. Left to right: Countdown game (stack size), Countdown game (range of numbers), Blocks World (number of blocks). Y-axis in log scale. Plots show NLRL takes approximately 10 times longer per episode compared to DERRL. See additional plots in Appendix~\ref{appendix:computation_time_comparison}.}
    \label{figure:run_time_comparison}
\vspace{-5pt}
\end{figure}

\textbf{Computational Complexity.} NLRL assigns trainable weights to all possible rules with a body size of 2, while DERRL allocates weights to each atom in the rule body. Given, $m$ atoms from $\mathrm{P}_{\text{ext}}$ and $V$, NLRL has C(m,2) learnable weights, while DERRL has 2m. Therefore, the training reduces from learning the best set of rules (in NLRL) to learning the best membership of the rules (in DERRL), leading to a lower computation time in DERRL. The computation time per episode is reduced by a factor of $\approx 10$ (Figure~\ref{figure:run_time_comparison}).
\\\textbf{Comparing Learned Rules.}
    NLRL learned rules in blocksworld are:
    \begin{align*}
    &move(X,Y) \leftarrow top(X), pred(X,Y); \; \; \;  move(X,Y) \leftarrow top(X), goal\_on(X,Y)\\
    &pred(X,Y) \leftarrow isFloor(Y), pred2(X); \;\;\;  pred2(X) \leftarrow on(X,Y), on(Y,Z)
    \end{align*}
With invented predicates $pred$ and $pred2$, this plan differs from DERRL in that the second rule doesn't verify if both X and Y are movable, failing to solve configurations where block b is below block a.
\textbf{Size of hypothesis space.} DERRL restricts the hypothesis space through the use of semantic constraints in the optimization problem, whereas large hypothesis space in NLRL limits its convergence in DoorKey and Traffic domains. The convergence in DERRL is slower without semantic constraints (see Appendix~\ref{appendix:computation_time_comparison} for ablations on DERRL with and without semantic constraints). \textbf{Expressiveness.}  NLRL can learn recursive rules by using templates as in meta-interpretive learning and predicate invention. While this is expressive, it can be hard to master. In contrast, DERRL learns non-recursive Datalog as Quinlan's FOIL \cite{quinlan_foil} but combines it with constraints that can be recursive to rule out redundancies.

\section{Conclusion}
\label{section:conclusion}
We proposed a neuro-symbolic approach to learn interpretable policies that are also generalizable. The representations that DERRL and RRL use are very similar to those used in the planning community.  We also significantly improve upon the scalability of existing state-of-the-art NLRL. Upgrading the approach to enable automatic learning of the required number of rules can be a potential research direction. Also, as a part of future work, it will be interesting to explore ways in which the proposed approach can be scaled to real-life applications requiring the need to process raw sensory inputs. 

\section*{Acknowledgement}
This work was partially supported by the Wallenberg AI, Autonomous Systems and Software Program (WASP) funded by the Knut and Alice Wallenberg Foundation.


%
%
%
\bibliographystyle{splncs04}
\bibliography{biblio}
%




\newpage

\begin{center}
\textbf{\Large Appendix for ``Deep Explainable Relational Reinforcement Learning: A Neuro-Symbolic Approach"}
\end{center}


\section{Semantic Constraints}
\label{appendix:semantic_constraints}
Following are the semantic constraints used for each setup
\subsubsection{The Countdown Game.}
\begin{align*}
&false \leftarrow goal(X), goal(Y)
&false \leftarrow less(X,Y), less(Y,X)\\
&less(X,Z) \leftarrow less(X,Y), less(Y,Z)
&false \leftarrow acc(X), acc(Y)\\
&false \leftarrow curr(X), curr(Y)
\end{align*}

\subsubsection{Blocks World Manipulation.}
\begin{align*}
&false \leftarrow isFloor(X), isFloor(Y)
&false \leftarrow on(X,Y), on(Y,X)\\
&false \leftarrow on(X,Y), on(X,Z)
&false \leftarrow top(X), on(Y,X)\\
&false \leftarrow top(Y), isFloor(Y)
&false \leftarrow goal(X,Y), goal(Y,X)
\end{align*}

\subsubsection{DoorKey Minigrid.}
\begin{align*}
&false \leftarrow carrying(X), notcarrying\\
&samecolor(X,Y) \leftarrow samecolor(Y,Z), samecolor(Z,X)\\
&samecolor(X,Y) \leftarrow samecolor(Y,X)\\
&false \leftarrow carrying(X), carrying(Y)\\
&false \leftarrow locked(X), unlocked
\end{align*}

\subsubsection{Traffic}
\begin{align*}
&between(X,Y,Z) \leftarrow between(X,Z,Y)\\
&false \leftarrow highest(X), highest(Y), highest(Z)
\end{align*}

\subsubsection{Gridworld Game}
\begin{align*}
&false \leftarrow curr(X,Y), curr(Y,X)
&false \leftarrow succ(X,Y), succ(Y,X)
\end{align*}



\section{Computation Time comparison}
\label{appendix:computation_time_comparison}
\begin{figure}[t]
	\centering
        \includegraphics[width=\linewidth]{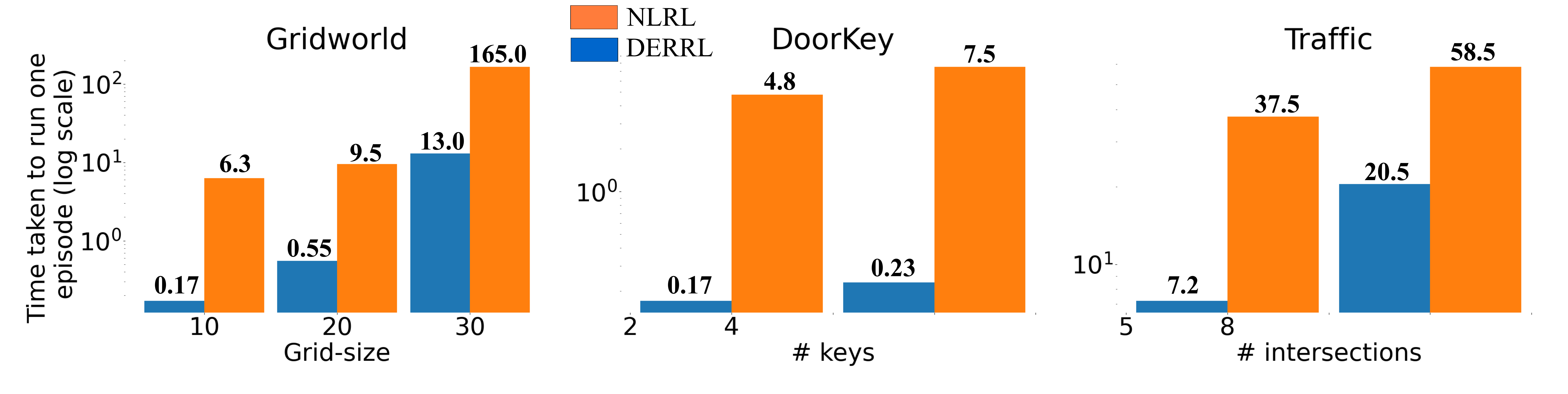}
		\caption{Comparison of run-time per episode of DERRL and NLRL with problem scaling: From left tor right: Gridworld (grid-size), DoorKey (number of keys), Traffic (number of intersections). The Y-axis is represented in log scale. The plots show that time taken per episode is significantly higher for NLRL, compared to DERRL.}
    \label{figure:appendix_run_time_comparison}
\end{figure}

\begin{figure}[t]
	\centering
        \includegraphics[width=\linewidth]{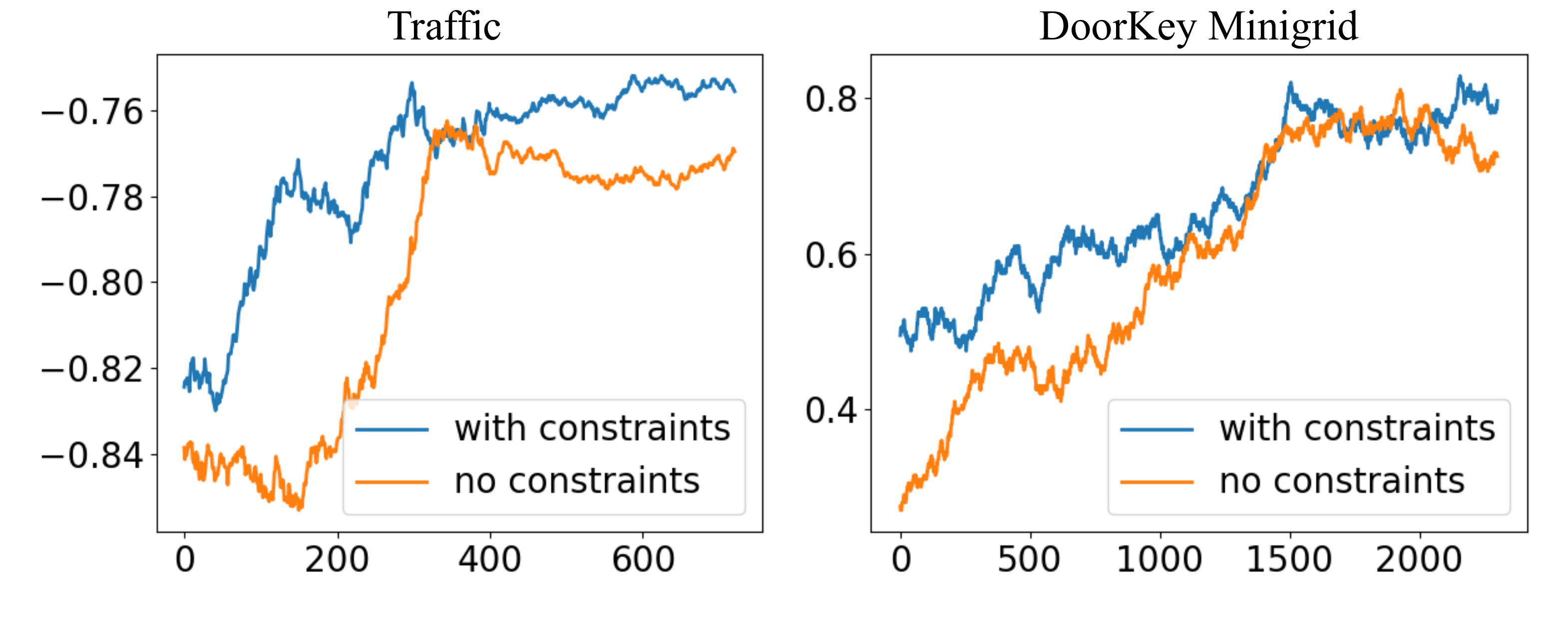}
        \caption{Ablations on semantic constraints. Figure shows how the use of semantic constraints lead to faster convergence in the Traffic (Left) and the DoorKey Minigrid (Right) domains.}
\label{figure:constraint-ablations}
\end{figure}

See Figure~\ref{figure:appendix_run_time_comparison} for run-time comparison of DERRL and NLRL for Gridworld, DoorKey, and Traffic domains. Additionally, as shown in Figure~\ref{figure:constraint-ablations}, without the use of semantic constraints, the convergence in DERRL is slower and the learned rules are redundant.

\section{Analogies with Program Synthesis}
\label{appendix:analogies_with_program_synthesis_cont}
\begin{figure*}
	\centering
		\includegraphics[width=\linewidth]{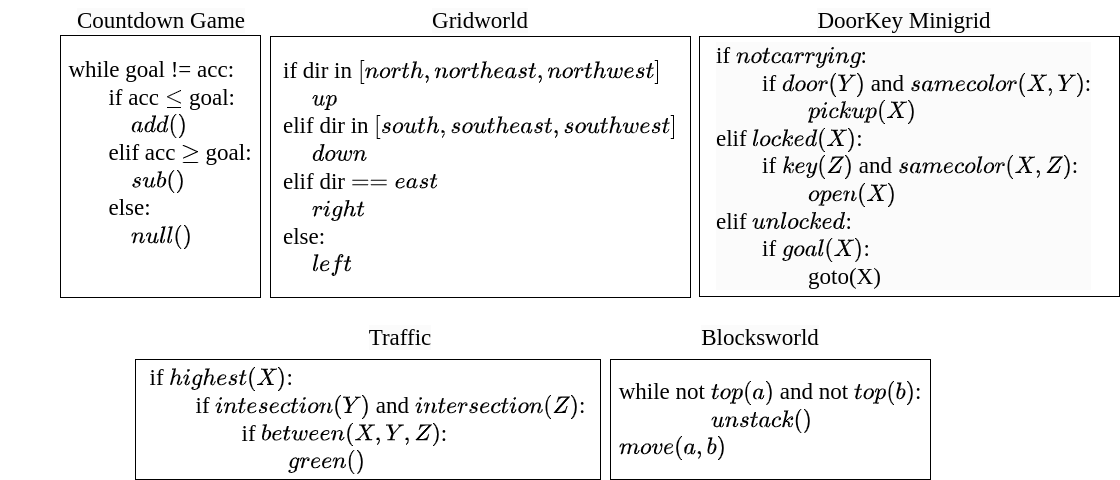}
		\caption{Programs for different environments: [Row 1]: from left to right, Countdown Game, Gridworld, DoorKey Minigrid. [Row 2]: from left to right, Traffic and Blocksworld.}
    \label{figure:programs}
\end{figure*}

In many ways, program synthesis techniques exhibit the same challenges as that of interpretable policy learning, wherein, programs generated using neural methods cannot generalize to new problems~\cite{graves2014neural,singh2017ap}, and symbolic methods employing search-based methods are faced with the curse of dimensionality~\cite{10.5555/1714168,10.1145/2451116.2451150,10.1145/2499370.2462174}. Our exploratory study indicates that our rule generation framework is closely related to the paradigm of rule-based program synthesis~\cite{10.1145/2240236.2240260,6679385}. The specification here is -- one needs to find a sequence of instructions (a program or a policy) that transforms the given state into a target state. However, in program synthesis, the instructions are usually deterministic, while in RRL they can be stochastic. Note, that DERRL selects at every time step the next action and moves to the next position and iterates.  Although DERRL uses non-recursive rules, the learned policies are applied iteratively, which results in programs as shown in Figure~\ref{figure:programs}.

\section{Fuzzy Conjunction Operators}
\label{appendix:fuzzy_conjunction_operators}
A fuzzy conjunction operator $* : [0,1]^e \mapsto [0,1]$ must satisfy the following conditions on a t-norm~\cite{ESTEVA2001271} (here, $e$ is the number of atoms in the rule body): 
\begin{itemize}
    \item commutativity: $a \ast b = b \ast a$
    \item associativity: $(a \ast b) \ast c = a \ast (b \ast c)$
    \item monotonicity:
        \begin{enumerate}
            \item $a_1 \leq a_2$ implies $a_1 \ast b \leq a_2 \ast b$
            \item $b_1 \leq b_2$ implies $a \ast b_1 \leq a \ast b_2$
        \end{enumerate}
    \item unit:
        \begin{enumerate}
            \item $a \ast 1 = a$
            \item $a \ast 0 = 0$
        \end{enumerate}
\end{itemize}
The operators that satisfy the aforementioned conditions are:
\begin{itemize}
    \item Godel t-norm: $a \ast b := \Min(a,b)$
    \item Lukasiewicz t-norm $\top_{\text{Luk}}(a,b) := \Max\{0, a+b-1\}$
    \item Product t-norm: $a \ast b := a \cdot b$ (ordinary product of real numbers)
\end{itemize}

Lukasiewicz t-norm was experimentally found to perform better than other alternatives.
\prop{} More generally, given a vector $\bm{y} \in [0, 1]^n$, Lukasiewicz t-norm 
\begin{equation}
\label{equation:general_t_norm}
    \top_{\text{Luk}}\bm{y} := \Max(0, \langle \bm{y}, \mathbb{1} \rangle - n + 1)
\end{equation}

\begin{proof}.
We prove it using mathematical induction.\\
\underline{Base Case:} For $\bm{y} = [a,b]$, eq.~\ref{equation:general_t_norm} is clearly true.

\begin{equation*}
    \top_{\text{Luk}}(a, b) := \Max(0, a + b - 2 +1)
\end{equation*}\\
\underline{Inductive Step:} Let, $\top_{\text{Luk}}\bm{y}$ hold for $\bm{y} \in [0, 1]^k$. We denote it as,
\begin{equation}
    \top_{\text{Luk}}\bm{y} := \Max(0, \langle \bm{y}, \mathbb{1} \rangle - k + 1)
\end{equation}\\
Now let $\tilde{y} \in [0, 1]^{k+1}$. Using the associativity property:

\begin{align*}
    \top_{\text{Luk}}\tilde{\bm{y}} &:= \Max(0, \top_{\text{Luk}}\bm{y} + y_{k+1} - 1) \\
    &:= \Max(0, \Max(0, \langle \bm{y}, \mathbb{1} \rangle - k + 1) + y_{k+1} - 1)
\end{align*}

\begin{itemize}
    \item Case 1:  $\langle \bm{y}, \mathbb{1} \rangle > (k - 1)$. It follows that $\Max(0, \langle \bm{y}, \mathbb{1} \rangle - k + 1) = \langle \bm{y}, \mathbb{1} \rangle - k + 1$
    \begin{align*}
        \top_{\text{Luk}}\tilde{\bm{y}} &:= \Max(0, \langle \bm{y}, \mathbb{1} \rangle - k + 1 + y_{k+1} - 1)\\
        &:= \Max(0, y_1 + \dots + y_k - k + 1 + y_{k+1} - 1)\\
        &:= \Max(0, \langle \tilde{\bm{y}}, \mathbb{1} \rangle - (k + 1) - 1)
    \end{align*}
    
    Thus, eqn.~\ref{equation:general_t_norm} is true.
    
    \item Case 2:  $\langle \bm{y}, \mathbb{1} \rangle \leq (k - 1)$. It follows that $\Max(0, \langle \bm{y}, \mathbb{1} \rangle - k + 1) = 0$
    \begin{align*}
        \top_{\text{Luk}}\tilde{\bm{y}} &:= \Max(0, y_{k+1} - 1)\\
        &:= 0  && \text{since } y_{k+1} \in [0, 1]
    \end{align*}
    
    Also, $\langle \bm{y}, \mathbb{1} \rangle \leq (k - 1) \implies \langle \tilde{\bm{y}}, \mathbb{1} \rangle \leq k$ since $y_{k+1} \in [0, 1]$.\\ $\therefore$ from eqn.~\ref{equation:general_t_norm}, $\top_{\text{Luk}}\tilde{\bm{y}} := 0$
\end{itemize}
\end{proof}

\section{Code}
All models were run on CPU nodes. The codes will be made available soon.

\end{document}